\documentclass[11pt,a4paper]{article}

\usepackage[a4paper, total={170mm,257mm}, left=20mm, top=20mm]{geometry}
\usepackage{newtxtext}
\usepackage{newtxmath}
\usepackage{graphicx}
\usepackage{siunitx}
\usepackage{float}
\usepackage{caption}
\usepackage{subcaption}
\usepackage{titlesec}
\usepackage{fancyhdr}
\usepackage{enumitem}
\usepackage{booktabs}
\usepackage{tabularx}
\usepackage{multicol}
\usepackage[T1]{fontenc}
\usepackage[colorlinks=true, linkcolor=blue!50!black, citecolor=blue!50!black, urlcolor=blue!50!black]{hyperref}

\titleformat{\section}
  {\bfseries\LARGE}
  {\thesection.}{0.8em}{}

\titleformat{\subsection}
  {\bfseries\Large}
  {\thesubsection}{0.6em}{}

\titleformat{\subsubsection}
  {\bfseries\itshape\normalsize}
  {\thesubsubsection}{0.6em}{}

\titlespacing*{\section}{0pt}{1.2em plus 0.5em minus 0.3em}{0.8em}
\titlespacing*{\subsection}{0pt}{1em plus 0.3em minus 0.2em}{0.5em}
\titlespacing*{\subsubsection}{0pt}{0.8em plus 0.3em minus 0.2em}{0em}

\title{\vspace{-1cm}\textbf{Technical Report of Mobile Manipulator Robot for Industrial Environments}}

\author{
\begin{minipage}[t]{0.95\textwidth}\centering
Aria Dadnavi,
Matin Jamshidi,
Kian Khatibi,
Shahriar Karimi Makhsous,
Saina Najafi Shabestari,
Seyed Sobhan Hosseini Hajibekandeh,
Zahra Mahaam,
Reyhane Mohasseli,
Ghazal Ghoreishi,
Amirhossein Gholizadeh Behbahani,
Mohammad Hossein Moaiyeri\\[1em]
\textit{Team Auriga, Faculty of Electrical Engineering, Shahid Beheshti University, Tehran, Iran}
\end{minipage}
}
\date{November, 2026}

\begin{document}
\maketitle
\thispagestyle{empty}

\vspace{-1.2cm}
\begin{center}
\small
\texttt{
\href{mailto:h_moaiyeri@sbu.ac.ir}{h\_moaiyeri@sbu.ac.ir}, \\
\{\href{mailto:a.dadnavi@mail.sbu.ac.ir}{a.dadnavi},
\href{mailto:mat.jamshidi@mail.sbu.ac.ir}{mat.jamshidi},
\href{mailto:k.khatibi@mail.sbu.ac.ir}{k.khatibi},
\href{mailto:s.najafishabestari@mail.sbu.ac.ir}{s.najafishabestari},
\href{mailto:z.maham@mail.sbu.ac.ir}{z.maham},
\href{mailto:r.mohasseli@mail.sbu.ac.ir}{r.mohasseli},
\href{mailto:g.ghoreishi@mail.sbu.ac.ir}{g.ghoreishi},
\href{mailto:am.gholizadeh@mail.sbu.ac.ir}{am.gholizadeh}\}@mail.sbu.ac.ir, \{\href{mailto:shahriar.karimi.m@gmail.com}{shahriar.karimi.m}, \href{mailto:sobhanhosseinih@gmail.com}{sobhanhosseinih}\}@gmail.com
}
\end{center}

\thispagestyle{empty}

\begin{abstract}
\noindent This paper presents the design and implementation of Auriga Robotics' mobile manipulator robot, developed at the Robotics and Intelligent Automation Laboratory, Department of Electrical Engineering, Shahid Beheshti University. Hereafter, this platform is referred to as the \textbf{@Work Robot}. The robot targets industrial tasks, with an emphasis on improving efficiency in repetitive or hazardous work environments. It uses a 4-wheel Mecanum drive for omnidirectional movement and a 5-degree-of-freedom manipulator arm with a 3D-printed gripper for object handling and navigation. The electronics consist of custom PCB boards built around ESP32 microcontrollers, together with an NVIDIA Jetson Orin NX that handles high-level processing and real-time system control. On the software side, Hector SLAM provides mapping and localization, A* handles path planning, and YOLO performs real-time object detection. These modules draw on fused sensor data to deliver robust navigation and reliable collision avoidance. Overall, the system shows effective hardware-software integration, with consistent performance in autonomous navigation, object handling, and task execution in industrial-like settings.

\vspace{0.5em}
\noindent\textbf{Keywords —} Industrial Robotics, Autonomous Navigation, Object Detection, Real-Time Control, Computer Vision
\end{abstract}

\newpage
\small
\tableofcontents
\newpage

\section{Introduction}

Robotics has become a core part of modern industrial settings, offering practical ways to optimize tasks and boost productivity. Within this context, @Work Robots work alongside human operators, helping create safer and more efficient workspaces. These robots can be deployed for a range of tasks, including production line support, transport in constrained spaces, object handling, and jobs that are difficult or unpleasant for humans to perform. This growing range of applications reflects the increasing importance of autonomous mobile manipulation in the future of industrial automation.

The goal of this project is to build a modular, highly adaptable platform that addresses these operational challenges — a versatile testbed that bridges academic research in mobile manipulation with practical, scalable industrial deployment.

\begin{figure}[H]
    \centering
    \includegraphics[width=\linewidth]{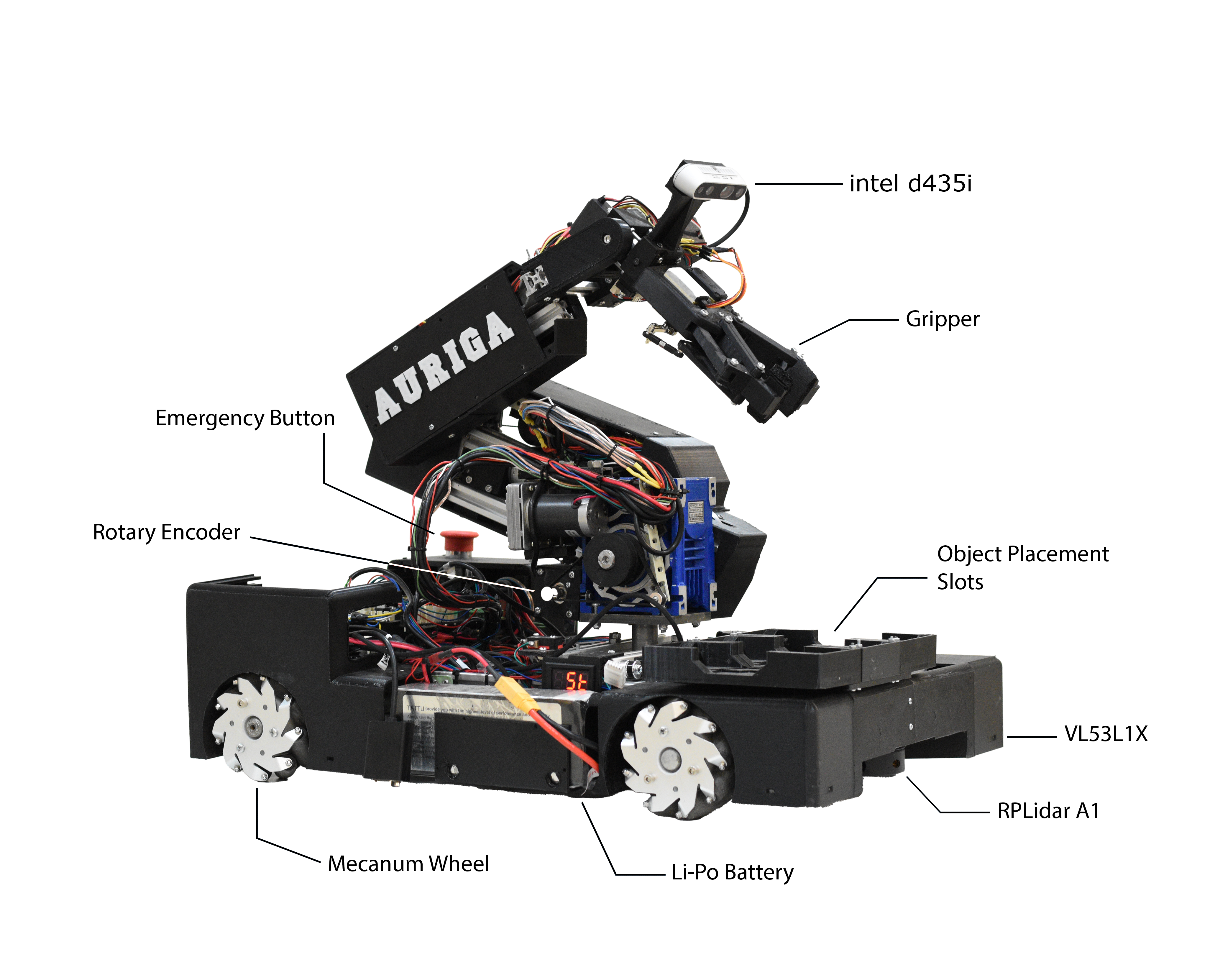}
    \caption{Auriga's @Work Robot.}
    \label{fig:robotpic}
\end{figure}

Auriga Robotics at Shahid Beheshti University developed an autonomous mobile manipulation platform for the RoboCup @Work league, shown in Figure~\ref{fig:robotpic}. The robot is built to handle perception, navigation, manipulation, and real-time decision-making. The project is part of the team's ongoing research in intelligent robotics and autonomous systems, aimed at producing a modular, extensible platform for academic research and experimental work in industrial automation. Among the main contributions of this work are custom electronics for optimized power and sensor integration, and a vision-based dynamic obstacle mapping framework that improves navigation in complex environments.

This paper covers the design process and system integration of the Auriga Robotics @Work platform, with emphasis on its modular hardware and intelligent control architecture. It is organized as follows: Section~2 presents the hardware design, including the mechanical structure, electronics, and sensors. Section~3 describes the embedded firmware architecture — sensor integration and communication between the onboard processing units — and details the software infrastructure and ROS-based communication architecture. Section~4 discusses perception and computer vision algorithms. Section~5 covers robot control, navigation, and decision-making.

\section{Hardware Description}

\subsection{Mechanical Design}

\subsubsection{Base Structure Design}
The chassis and base structure are built mainly from aluminum profiles, giving a frame that is both lightweight and sturdy. The base measures 65~cm in length, 42~cm in width, and 56~cm in height, with a total weight of roughly 17~kg. Several 3D-printed components and plates were also added to the structure for a more integrated, customized design. Locomotion relies on four Mecanum wheels, each driven independently by a DC motor, giving the robot omnidirectional movement and high maneuverability. To cut down on task execution time, the object-handling area sits at the front of the robot, so the manipulator never has to rotate backward during pick-and-place operations — a choice that meaningfully improves efficiency and shortens task duration.

\subsubsection{Manipulator Design}
\begin{figure}[H]
    \centering
    \begin{subfigure}{0.45\textwidth}
        \centering
        \includegraphics[height=6cm]{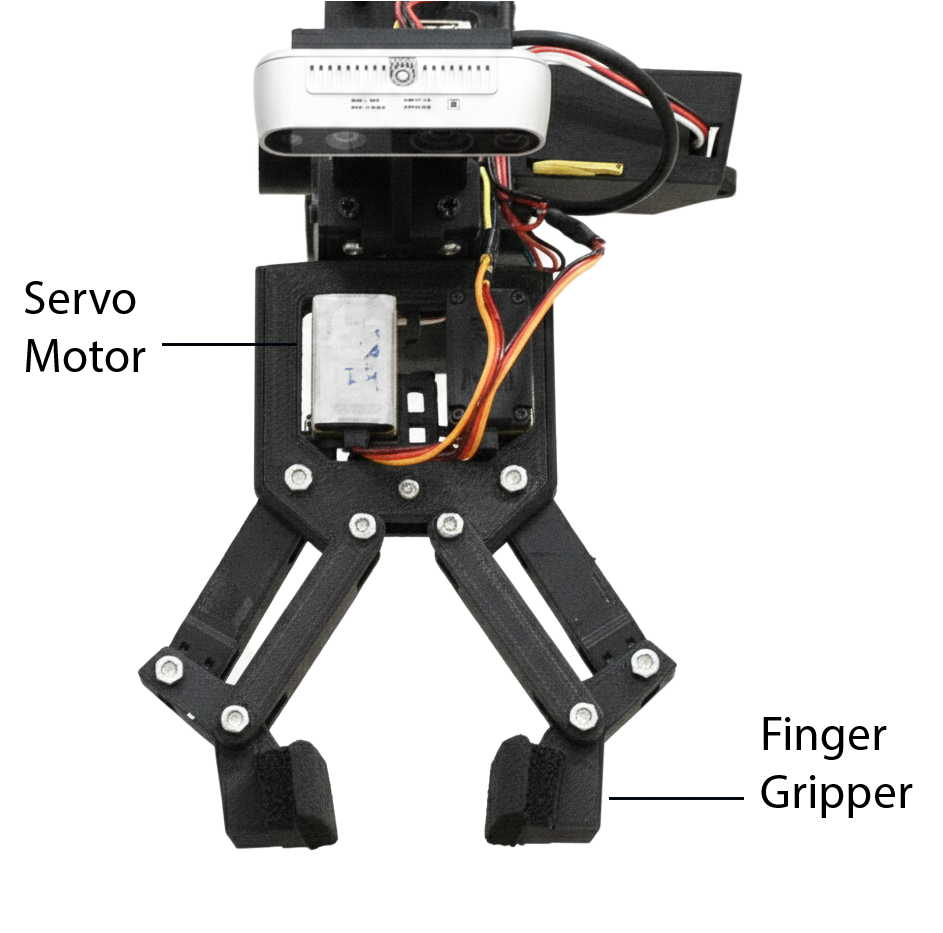}
        \caption{}
        \label{fig:prototype}
    \end{subfigure}
    \hfill
    \begin{subfigure}{0.45\textwidth}
        \centering
        \includegraphics[height=6cm]{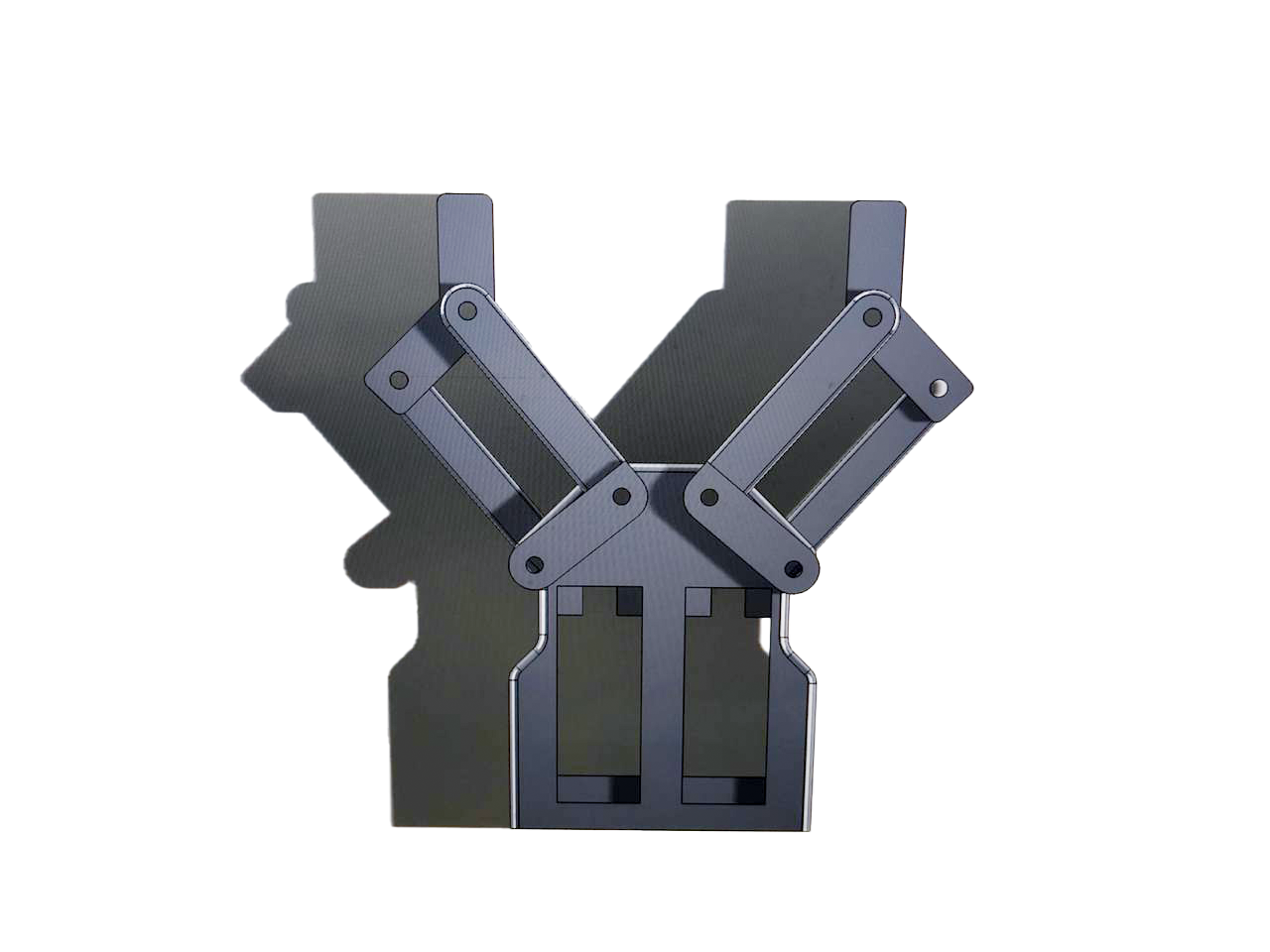}
        \caption{}
        \label{fig:model}
    \end{subfigure}

    \caption{Design and implementation of the robotic gripper:
    (a) Fabricated prototype;
    (b) 3D model.}
    \label{fig:gripper}
\end{figure}

The manipulator arm has 5 joints and 3 links, ending in a purpose-built gripper, shown in Figure~\ref{fig:gripper}. It reaches up to 65~cm and can carry a payload of roughly 450~g. The first three joints provide rotational motion about the robot's principal axes, driven by three worm gearboxes powered by DC motors that supply enough torque and positioning precision. The fourth and fifth joints control gripper orientation using servo motors, letting the manipulator hold objects at different angles. Two more servo motors actuate the gripper fingers for grasping and releasing.

\subsection{Electrical Design}

Auriga's @Work platform relies on two main custom electronic boards: one controls the manipulator arm, the other manages base movement. Keeping them separate allows the arm and base systems to be tested and debugged independently, which simplifies development and maintenance. The system also includes custom ESP32~\cite{esp32} boards and a support board with op-amps and other ICs to improve the efficiency of the main boards, covered in detail below. An emergency button is wired into the main board for operational safety. Both main boards draw power from a four-cell 16000~mAh lithium-polymer battery, with voltage regulators supplying the correct operating voltage to each circuit.

\subsubsection{Body Main Control Board}
\begin{figure}[H]
    \centering
    \begin{subfigure}{0.8\textwidth}
        \centering
        \includegraphics[width=0.5\linewidth]{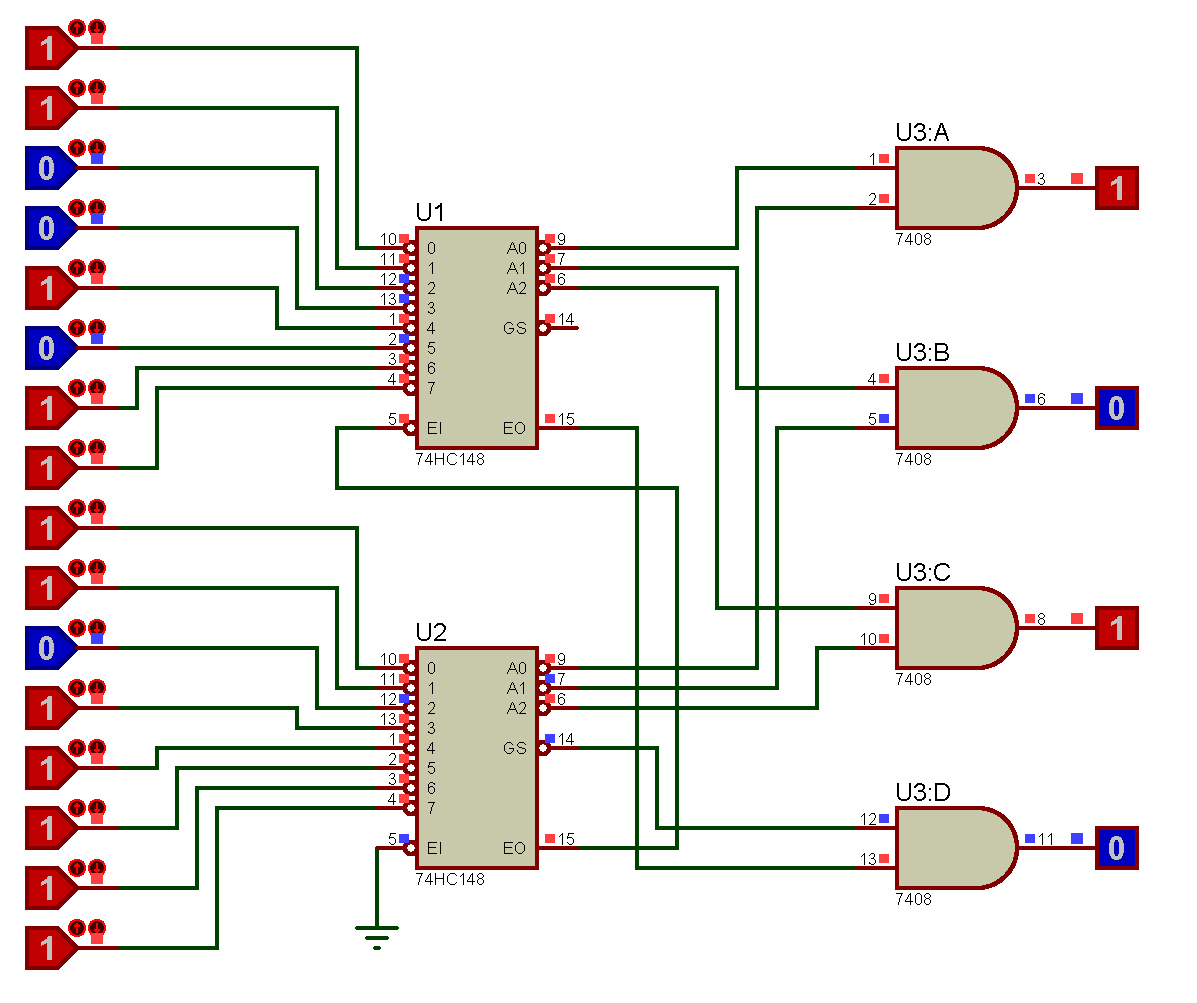}
        \caption{}
        \label{fig:sim}
    \end{subfigure}

    \vspace{0.5em}

    \begin{subfigure}{0.47\textwidth}
        \centering
        \includegraphics[height=6cm]{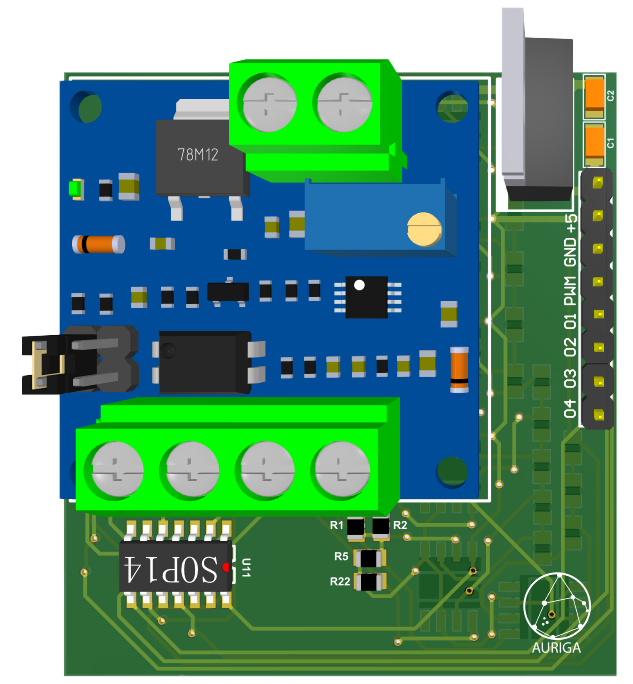}
        \caption{}
        \label{fig:pcbtop}
    \end{subfigure}
    \hfill
    \begin{subfigure}{0.47\textwidth}
        \centering
        \includegraphics[height=6cm]{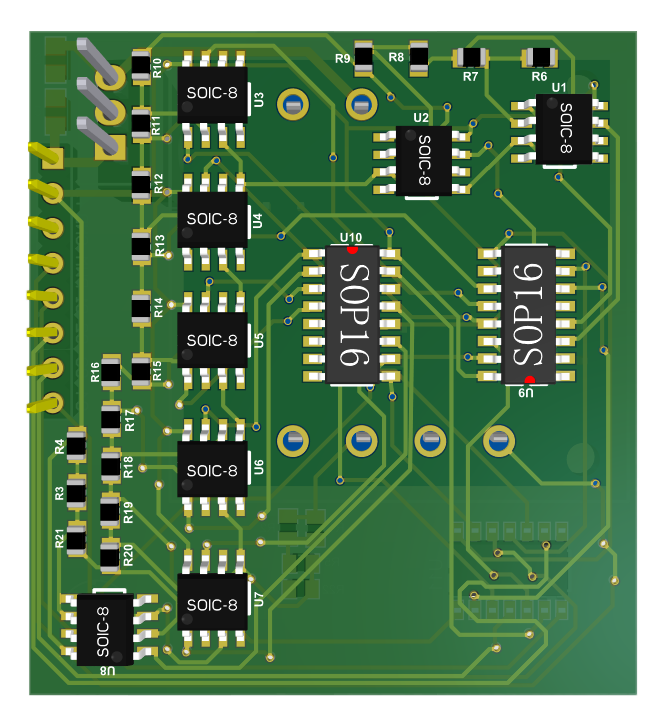}
        \caption{}
        \label{fig:pcbbottom}
    \end{subfigure}

    \caption{PWM-to-binary conversion system:
    (a) Simulation environment demonstrating the proposed method,
    (b) Top layer of the fabricated PCB,
    (c) Bottom layer of the fabricated PCB.}
    \label{fig:pwmbinary}
\end{figure}

\begin{figure}[H]
    \centering
    \includegraphics[width=0.8\linewidth]{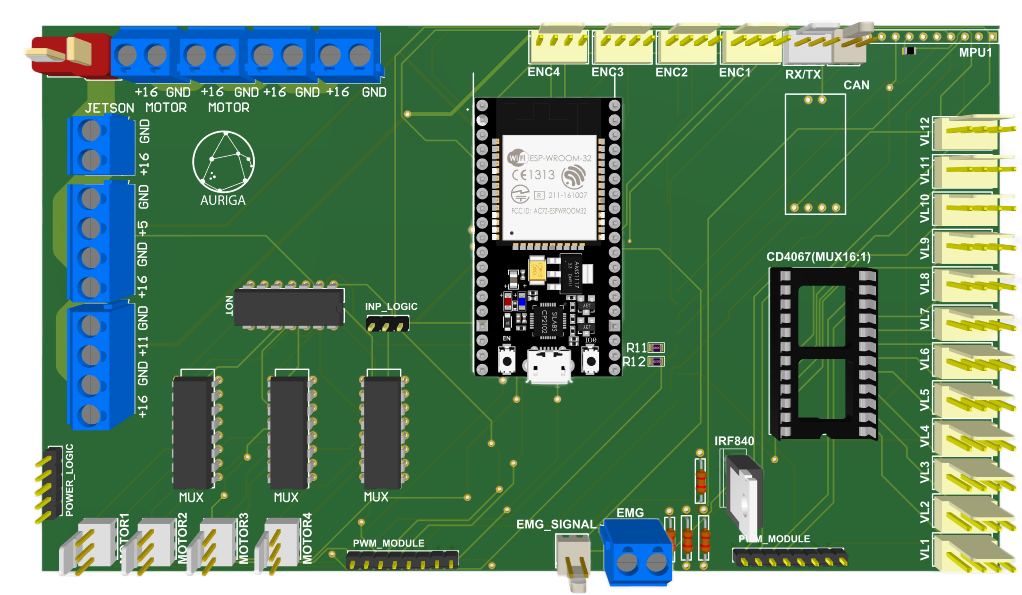}
    \caption{Main control board PCB.}
    \label{fig:mainboard}
\end{figure}

On this board, shown in Figure~\ref{fig:mainboard}, LM2596 switching regulators supply several voltage levels: 11~V for the encoders, 12~V for the NVIDIA Jetson Orin NX, and 5~V for the ICs, microcontrollers, range-measurement sensors, gyroscopes, and CAN module. The gyroscopes and range-measurement sensors talk to the microcontroller over I\textsuperscript{2}C, while data between the microcontroller and the Jetson Orin NX moves over a UART serial connection. The DC motors are driven by the HW-039 motor driver, rated for 6--27~V DC and a continuous current of 43~A, which suits high-power applications well. We also implemented a dedicated method for motor direction control that uses a single ESP32 pin to manage the direction of all four wheel motors; its functional simulation appears in Figure~\ref{fig:pwmbinary}a, and the fabricated PCB (top and bottom layers) is shown in Figures~\ref{fig:pwmbinary}b and~\ref{fig:pwmbinary}c. The PWM signal from the ESP32 is first converted into an analog voltage, then digitized using op-amps and a resistor-divider network to set up a stable reference. Two 74LS148 encoder ICs extract a 4-bit value from the resulting 16-bit range, giving up to 16 distinct directional states for the motors. These bits then pass through a 7404 NOT IC and a MUX/DeMUX IC to set the motor directions. The same approach is used to read data from the VL53L1X sensors through a MUX.

The emergency button on the body board can safely disable and re-enable the motors. Its safety circuit uses an IRF840 MOSFET and relays to reliably cut power when needed. Circuit design, PCB layout in Altium Designer, and hardware assembly were all carried out in-house by the electronics team.

\subsubsection{Arm's Board}
\begin{figure}[H]
    \centering
    \includegraphics[width=0.85\linewidth]{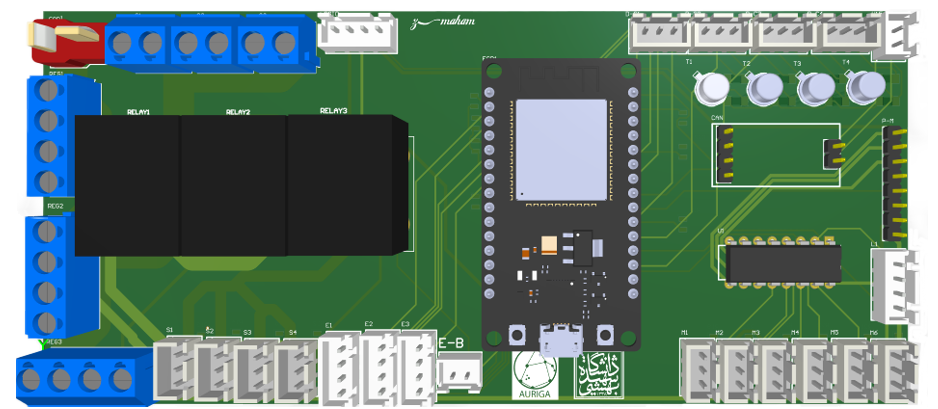}
    \caption{Arm's board PCB.}
    \label{fig:armboard}
\end{figure}

\begin{figure}[H]
    \centering
    \includegraphics[width=0.3\linewidth]{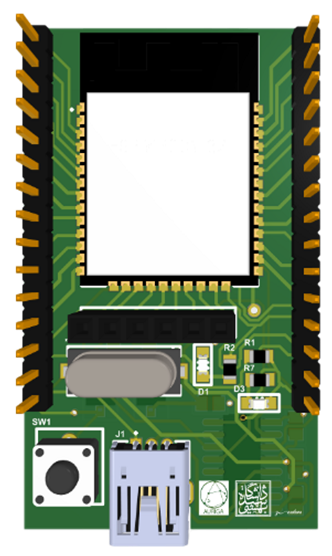}
    \caption{Custom-designed ESP32 PCB.}
    \label{fig:esp32board}
\end{figure}

On this board, shown in Figure~\ref{fig:armboard}, three LM2596 switching regulators supply different voltage levels: 8~V for one servo motor and three encoders, 6~V for three servo motors, and 5~V for the microcontroller and logic ICs. An AMS1117 regulator supplies an additional 3.3~V logic voltage for the DC motor driver. The board is built around a custom ESP32-based control unit, shown in Figure~\ref{fig:esp32board}. Motor actuation on the arm is handled by two HW-039 drivers and one L298N driver. The board also includes three relays for the emergency button system, four BJT transistors for switching, and a CAN module for communication with the main body board. Four servo motors are used here as well, driven using the same PWM-to-binary conversion method described above. To save ESP32 GPIO pins, an 8:1 analog multiplexer reads the microswitch signals used as reference inputs for motion initialization.

\subsubsection{Sensors}

\paragraph{Intel RealSense D435i Depth Camera.} The Intel RealSense D435i~\cite{realsense_d435i} is an RGB-D camera that handles depth perception and visual data acquisition, feeding synchronized RGB and depth data to the robot's main processor for perception and scene understanding.

\paragraph{RPLIDAR A1.} The RPLIDAR A1 is a time-of-flight LiDAR that performs a 360-degree laser scan to detect surrounding obstacles. Mounted at the front of the robot, it is used mainly for mapping, navigation, and physical obstacle detection.

\paragraph{Rotary Encoder.} We use the E38S6G5-100B-G24N rotary encoder, which measures shaft rotation to estimate angular displacement and velocity. It is used on both the arm and the mobile base to track joint positions, wheel velocities, and overall system state.

\paragraph{Range Sensor.} The VL53L1X~\cite{vl53l1x} is a time-of-flight distance sensor that covers LiDAR blind spots, improving collision avoidance during lateral and backward movement thanks to its speed and measurement accuracy.

\paragraph{Gyroscope.} Robot orientation is measured using both CMPS14 and MPU6050 sensors. Their outputs are fused for a more reliable angle estimate, which improves localization accuracy.

\paragraph{Load Cell.} A load cell in the gripper measures grip force, with a resolution of 0.5~grams and a threshold of 10~grams. A sudden drop in force signals possible slippage or loss of the object. It also enables controlled force application, which keeps stress on the servo motors low and extends their lifespan.

\section{Embedded Firmware, Communication, and ROS Architecture}

\subsection{Embedded Firmware and Design}

The platform relies on three main processing units: one NVIDIA Jetson Orin NX module and two ESP32 microcontrollers. This split is meant to improve processing efficiency, keep subsystems in sync, and simplify overall data management.

The NVIDIA Jetson Orin NX acts as the high-level computing platform, running the perception and control algorithms; its computational headroom lets it comfortably keep up with the system's processing demands.

Of the two embedded controllers, one sits in the lower chassis, where it manages communication between subsystems, handles data exchange, and gathers measurements from onboard sensors; in the system architecture this is the lower-level module referred to as the Navigation subsystem. The second ESP32 controller is dedicated to the robotic manipulator and is called the Arm module. It stays in contact with the Navigation subsystem, drives the arm actuators, and reads joint position feedback from the rotary encoders to keep motion under precise control.

\subsubsection{Embedded Firmware and Communication}
The embedded subsystem handles sensor data acquisition, deterministic communication between units, and real-time actuator control, with continuous, low-latency data exchange keeping the whole system in sync.

\begin{figure}[H]
    \centering
    \includegraphics[width=1\linewidth]{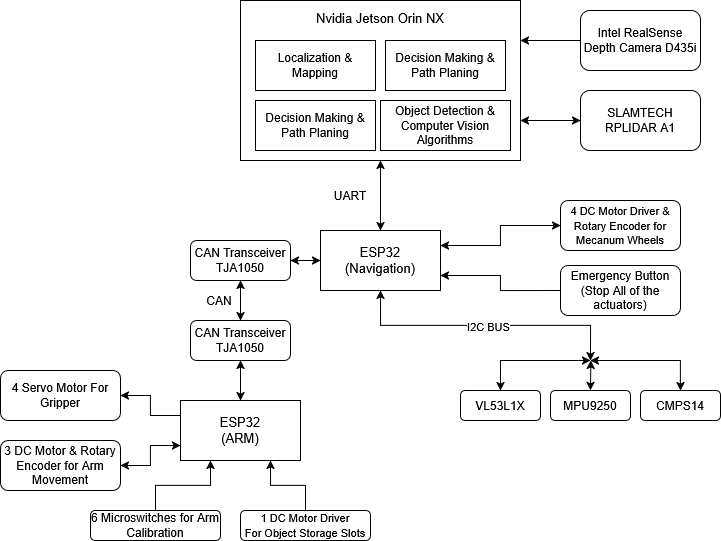}
    \caption{Hardware architecture and communication block diagram of the robotic system, illustrating the interconnection between the Nvidia Jetson, ESP32 controllers, sensors, and actuators.}
    \label{fig:embedded_architecture}
\end{figure}

The Navigation controller acts as the central communication hub in the embedded architecture. It gathers sensor measurements and receives status updates from the ARM unit, then sends this data to the Jetson module over UART at 115200~bps, as shown in Figure~\ref{fig:embedded_architecture}. In the other direction, high-level commands from the Jetson go to the Navigation unit, which either executes them directly or passes the relevant commands on to the ARM controller. Communication between the Navigation and Arm units runs over CAN, chosen for its resistance to electrical noise and suitability for reliable real-time embedded communication.

Both embedded controllers run FreeRTOS-based firmware to make efficient use of resources, avoid task blocking, and guarantee deterministic scheduling. On the Arm controller, one task handles PID-based motor control while a separate task manages communication with the Navigation unit and the data it exchanges. The Navigation controller follows a similar pattern: one task acquires and preprocesses sensor data before sending it to the Jetson, another manages communication with the Arm unit, and a third runs PID-based velocity control for the navigation motors.

\subsubsection{Sensor and Actuator Architecture}
A range of sensors provides situational awareness and monitors the robot's own state. For distance to obstacles, the LiDAR is supplemented with optical time-of-flight sensors (VL53L1X). Since these communicate over I\textsuperscript{2}C, a hardware multiplexer managed by the Navigation controller allows them to run simultaneously. To keep pin usage on the controller low, a PWM-to-binary counter module is used as well.

Inertial measurements come from the MPU6050 and CMPS14 sensors, which track the robot's angular acceleration and orientation and communicate directly with the Navigation unit.

Several microswitches handle manipulator calibration, with their states sent to the Arm controller. Rotary encoders measure the angular position and velocity of the robot's actuators, feeding the control loops.

All control computations run on the respective controllers, which then send the resulting commands to the motor drivers for execution.

\subsection{Robot Communication, Workflow Infrastructure and ROS Architecture}
\label{sec:atwork_ros}

This section covers the software infrastructure behind the @Work Robot's operation, from receiving a task order from the referee box down to executing low-level actuator commands on the embedded hardware. The system runs on the Robot Operating System (ROS) Noetic~\cite{ros} under Ubuntu~20.04 LTS, on an NVIDIA Jetson Orin NX onboard computer. The description below follows the system's data flow in the order it occurs during a competition run: task reception, decision-making, cross-machine communication, task execution and navigation, perception integration, and finally low-level actuation over UART. The internal logic of each subsystem — how a path is planned, how an object is detected, how a manipulation is performed — is covered in its own dedicated section elsewhere in this report; here we focus purely on the architectural glue connecting these subsystems.

\subsubsection{Task Reception and Multi-Master Communication}

\paragraph{Task Reception via atwork\_commander.} \label{subsec:commander_reception}
During a competition run, task orders come from the referee box via the \texttt{atwork\_commander}~\cite{atwork_commander} framework, which publishes a structured task description containing an ordered set of object manipulation subtasks, each with a source location, destination location, and object identifier. A dedicated \textit{decision-making} node (\texttt{decision\_making}) receives this description by subscribing to \texttt{/atwork\_commander/object\_task}. Once the task message arrives, the node unsubscribes from the topic, since only one task order is issued and processed per run.

\paragraph{Multi-Master Communication Bridge and atwork\_commander Integration.} \label{subsec:multimaster}
Task instructions for the @Work Robot come from the RoboCup @Work referee box
(\textit{refbox}), implemented as the \texttt{atwork-commander} ROS package. In the
competition setup, the refbox runs its own \texttt{roscore} on the referee's machine
(\textit{Machine~1}), while each competing robot runs an independent \texttt{roscore}
on its onboard computer (\textit{Machine~2}). Getting task messages from the refbox
to the robot, and robot state back in the other direction, therefore means bridging
two independent ROS masters.

The \texttt{atwork-commander} package originally used
\texttt{multimaster\_fkie}~\cite{multimaster_fkie} as its multi-master communication
backend. That library has since been deprecated and no longer works under
ROS~Noetic. Since no maintained drop-in replacement existed that preserved
compatibility with \texttt{atwork\_commander\_msgs} and the expected topic interface,
we designed and built a custom, lightweight bridge from scratch to restore full
interoperability between the refbox and the robot.

\subparagraph{System Architecture.}

\begin{figure}[H]
    \centering
    \includegraphics[width=\linewidth]{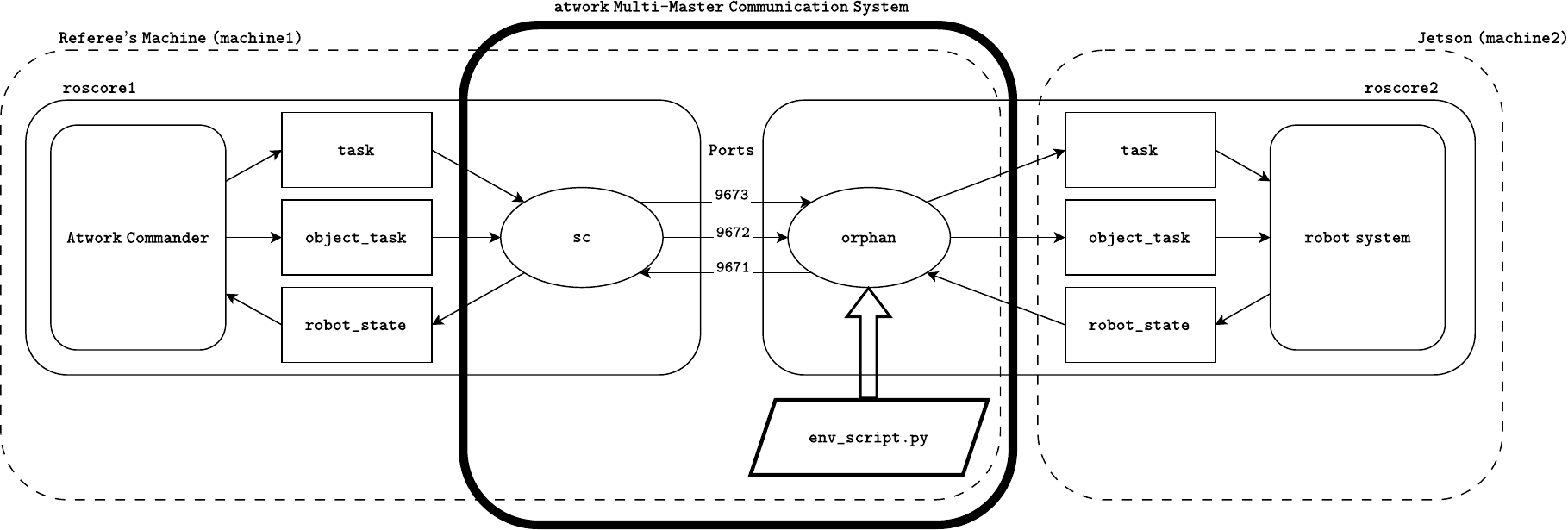}
    \caption{Multi-master ROS bridge architecture. All components reside on Machine~1.
    \texttt{sc} connects to the local atwork-commander roscore; \texttt{orphan} connects
    to the remote robot roscore on Machine~2. Topic data is relayed between them over
    Machine~1's loopback network. The \texttt{env\_script.py} supervisor manages the
    lifecycle of \texttt{orphan}, relaunching it against a new robot roscore on demand.}
    \label{fig:bridge_arch}
\end{figure}

All components of the bridge run on Machine~1 (the referee's machine). The system
consists of two C++ ROS nodes---\texttt{sc} and \texttt{orphan}---and a Python
supervisor script (\texttt{env\_script.py}); their relationships are shown in
Figure~\ref{fig:bridge_arch}.

\subparagraph{The \texttt{sc} node.}
The \texttt{sc} node registers as a ROS node under the atwork-commander's roscore on
Machine~1. It subscribes to the refbox's output topics---\texttt{/atwork\_commander/task}
and \texttt{/atwork\_commander/object\_task}---while also listening for incoming data
on a dedicated TCP port. When a message arrives on either subscribed topic, \texttt{sc}
serializes it with the standard \texttt{ros::serialization} API and forwards the binary
payload to the \texttt{orphan} node over Machine~1's loopback network. Going the other
way, \texttt{sc} receives serialized \texttt{RobotState} data from \texttt{orphan} on
its listening port, deserializes it, and republishes it to
\texttt{/atwork\_commander/robot\_state} on the atwork-commander roscore.

\subparagraph{The \texttt{orphan} node.}
The \texttt{orphan} node also runs on Machine~1, but registers under the
\emph{robot's} roscore on Machine~2 by launching with \texttt{ROS\_MASTER\_URI}
pointed at the robot's network address. This is the key design decision behind the
bridge: since a running ROS node cannot switch its hosting roscore after
initialization, \texttt{orphan} is started fresh for each robot connection with the
right master URI, so it looks like a native participant in the robot's ROS graph. Once
running, \texttt{orphan} receives the serialized payloads that \texttt{sc} forwards over
the loopback network, deserializes them back into their original message types, and
publishes them to the topics the robot expects: \texttt{/atwork\_commander/task} and
\texttt{/atwork\_commander/object\_task} on Machine~2's roscore. At the same time, it
subscribes to \texttt{/atwork\_commander/robot\_state} on the robot's roscore and
forwards any incoming \texttt{RobotState} messages back to \texttt{sc}, closing the
loop on the bidirectional bridge. Table~\ref{tab:bridge_topics} summarizes the three
topic streams and their corresponding TCP ports.

\begin{table}[H]
    \centering
    \small
    \caption{ROS topics bridged between the refbox roscore (Machine~1) and the robot roscore (Machine~2)}
    \renewcommand{\arraystretch}{1.2}
    \begin{tabular}{@{}llll@{}}
\toprule
\textbf{Topic} & \textbf{Message Type} & \textbf{Direction} & \textbf{TCP Port} \\ \midrule
\texttt{/atwork\_commander/task}         & \texttt{Task}       & Refbox $\rightarrow$ Robot & 9673 \\
\texttt{/atwork\_commander/object\_task} & \texttt{ObjectTask} & Refbox $\rightarrow$ Robot & 9672 \\
\texttt{/atwork\_commander/robot\_state} & \texttt{RobotState} & Robot $\rightarrow$ Refbox & 9671 \\
\bottomrule
\end{tabular}
\label{tab:bridge_topics}
\end{table}

\subparagraph{The \texttt{env\_script.py} supervisor.}
During competition, the referee has to send the same generated task to each competing
robot in turn. Since a node can't migrate between roscores at runtime, a fresh
instance of \texttt{orphan} has to be spawned for each robot connection. The
\texttt{env\_script.py} script automates this: it asks the referee for the
IP address of the target robot's roscore, polls for reachability with
\texttt{rosgraph.is\_master\_online()}, and once confirmed, launches \texttt{orphan}
as a subprocess with \texttt{ROS\_MASTER\_URI} set accordingly. It keeps monitoring the
subprocess, and if the robot roscore drops offline, the connection is
lost, or the referee disconnects manually, the script kills the current
\texttt{orphan} instance and asks for the next robot's IP, ready to repeat the
process for the next competitor.

\paragraph{Wire Protocol and Reliability.}
All three TCP channels use the same length-prefixed framing scheme. Each
transmitted unit consists of a 4-byte big-endian payload length followed immediately
by the ROS-serialized message bytes:
\[
\underbrace{L_3\;L_2\;L_1\;L_0}_{\text{4-byte length (network order)}}
\;\Big|\;
\underbrace{b_0\;b_1\;\cdots\;b_{N-1}}_{\text{ROS-serialized payload (\(N\;\)bytes)}}
\]
Receivers implement a two-state parser
(\textsc{read\_header} $\rightarrow$ \textsc{read\_payload}) that accumulates bytes
across successive \texttt{read()} calls before committing to deserialization. This
handles TCP stream fragmentation correctly regardless of MTU or scheduling jitter, so
no message is lost or corrupted in transit.

Each outbound TCP channel is backed by an \texttt{AsyncPortSender} object, which owns
a dedicated \texttt{std::thread} that drains a mutex-protected message queue over a
persistent TCP connection. This keeps ROS subscription callbacks decoupled from
network I/O, so transient link latency doesn't stall the ROS spin loop. Connections
are checked with periodic keep-alive probes; on failure, the sender tears down and
re-establishes the connection automatically, recovering from transient
network interruptions without a node restart.

\subsubsection{Task Execution Pipeline}

\paragraph{Decision-Making Node.}
\label{subsec:decision_node}

The \texttt{decision\_making} node takes the parsed task order and works out an optimized sequence of pick-and-deliver operations, factoring in station locations, object types, and estimated travel and manipulation times. We don't detail the internal optimization logic here, since it's not part of the ROS communication architecture. From an infrastructure standpoint, this node acts as a single-shot translator: it converts the \texttt{atwork\_commander} task representation into an internal, robot-specific one, and republishes the result once the full task order has been resolved.

The resulting task sequence is published on a custom topic, \texttt{decision}, using a nested custom message type (\texttt{estrings}, itself made up of a repeated string-array message \texttt{estring}) that encodes each subtask as a row of string-serialized fields (object identifier, task status, workstation type, and target pose). This message is consumed exactly once, by the navigation and task-execution node described in Section~\ref{subsec:main_node}, marking the handoff between the decision layer and the execution layer.

\paragraph{Navigation and Task-Execution Node.}
\label{subsec:main_node}

A single node, \texttt{main}, orchestrates the robot centrally. It subscribes to the \texttt{decision} topic and keeps an internal task queue built from the received message. This node runs a finite-state machine that governs the robot's high-level behavior — navigation, alignment, manipulation triggering, inter-task transitions — and it's the point where all other subsystems, including localization, perception, and low-level actuation, come together. Because this node touches every other subsystem in the architecture, its outward-facing interfaces are described individually in the following subsections, while its internal state-machine logic and control algorithms are covered elsewhere in this report.

\subsubsection{Sensor and Perception Interfacing}

\paragraph{Localization and Low-Level Sensor Interfacing.}
\label{subsec:esp_comms}
Low-level actuation and sensor acquisition are delegated to a dedicated microcontroller, reached through a Python bridge node, \texttt{esp\_comms}, which talks to the microcontroller over a UART serial link (\texttt{/dev/ttyTHS0}) using a fixed-format ASCII packet protocol. This node handles two jobs at once: serializing outgoing actuator commands into the packet format the microcontroller expects, and deserializing incoming sensor telemetry back into ROS messages.

On the outgoing side, \texttt{esp\_comms} subscribes to \texttt{motor4} (wheel actuation commands) and \texttt{arm\_datas} (manipulator commands), both published by \texttt{main}, and merges them into a single serial packet sent to the microcontroller each cycle. On the incoming side, it parses the microcontroller's serial reply and republishes the extracted data as three separate topics: \texttt{encoder4} (wheel encoder readings), \texttt{VL} (time-of-flight range array), and \texttt{cmps} (compass heading), all consumed by \texttt{main} for localization. The microcontroller's arm-response status isn't published as a topic at all; instead it's written directly to the ROS parameter server (\texttt{arm\_response}) and polled by \texttt{main} each control cycle. We use this parameter-server channel deliberately, as a lightweight, low-latency status flag rather than a queued topic, since only the most recent value ever matters to the state machine.

The \texttt{main} node fuses the encoder, compass, and range data internally to maintain the robot's global pose estimate; this fusion logic is covered in the localization section and isn't repeated here.

\paragraph{Perception Integration.}
\label{subsec:perception_integration}

Visual perception is handled by two cooperating nodes, \texttt{camera\_node} and \texttt{vision\_node}, both part of the \texttt{arm} package. \texttt{camera\_node} initializes and continuously polls an Intel RealSense D455 camera, and instead of publishing raw color and depth frames over ROS topics, it writes them directly into two POSIX shared-memory segments (\texttt{image\_frame} and \texttt{depth\_frame}). \texttt{vision\_node} maps the same shared-memory segments and reads the latest available frames directly, skipping ROS message serialization entirely for this high-bandwidth path. This mirrors the shared-memory pattern used elsewhere in the team's software stack for large image payloads.

Beyond camera-based perception, \texttt{vision\_node} also manages a 2D LiDAR unit (RPLidar) on its own, without a separate ROS driver node; the LiDAR is opened directly as a serial device from within \texttt{vision\_node} on a dedicated background thread, and its scan data is consumed internally by the same node rather than published on a topic. This keeps all mapping-relevant sensing — camera and LiDAR — inside a single perception process.

\texttt{vision\_node} and \texttt{main} coordinate through one topic plus the ROS parameter server. \texttt{vision\_node} publishes its detection output --- object, hole, or station-relative measurements, depending on the active mode --- on \texttt{/vision\_out} as a two-dimensional array packed into a single \texttt{Int32MultiArray} message with explicit layout dimensions. \texttt{main}, on the other hand, doesn't command the perception mode through a topic; instead it writes the desired operating mode and target object identifier, along with the robot's current global pose, to the parameter server (\texttt{mode}, \texttt{obj}, \texttt{x}, \texttt{y}, \texttt{theta}), which \texttt{vision\_node} polls every cycle. This keeps the perception node a passive follower of the task-execution state machine, with no dedicated command topic or service needed.

\subsubsection{System Integration and Summary}

\paragraph{System Bring-Up.}
\label{subsec:bringup}

The system comes up through two independent ROS launch files rather than one unified configuration. The first, \texttt{esp.launch}, starts the \texttt{esp\_comms} bridge and the \texttt{main} navigation/task-execution node together, reflecting how tightly they're coupled through the motor, encoder, and arm-command topics described in Section~\ref{subsec:esp_comms}. The second, \texttt{vision.launch}, starts \texttt{camera\_node} and \texttt{vision\_node} together, reflecting their shared-memory coupling from Section~\ref{subsec:perception_integration}. This split lets the perception subsystem and the navigation/actuation subsystem be started, restarted, and debugged independently during development and testing.

\paragraph{Summary of the Communication Architecture.}
\label{subsec:summary_architecture}

\begin{figure}[H]
    \centering
    \includegraphics[width=0.9\linewidth]{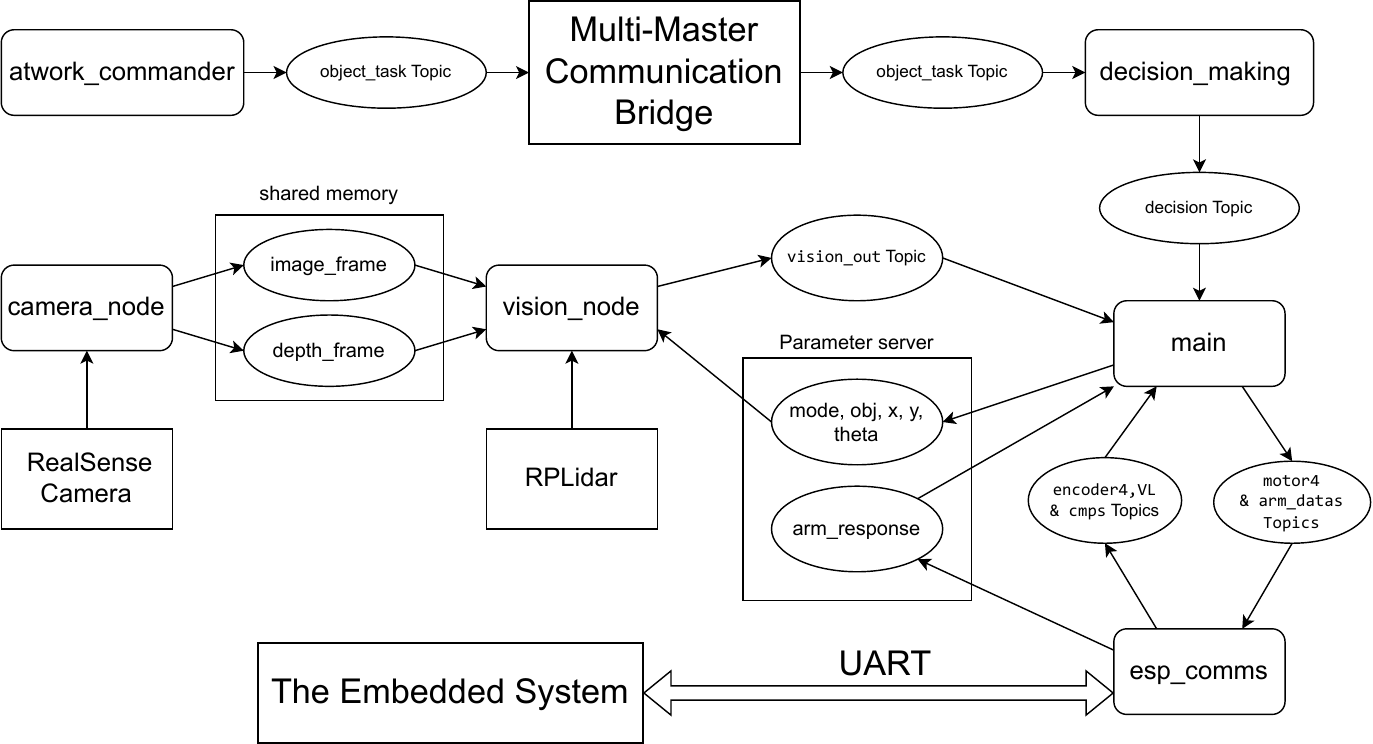}
    \caption{Nodes and communication graph of the @Work Robot's ROS architecture, showing publish--subscribe relationships between \texttt{decision\_making}, \texttt{main}, \texttt{esp\_comms}, \texttt{camera\_node}, and \texttt{vision\_node}.}
    \label{fig:atwork_rosgraph}
\end{figure}

\begin{figure}[H]
    \centering
    \includegraphics[width=0.95\linewidth]{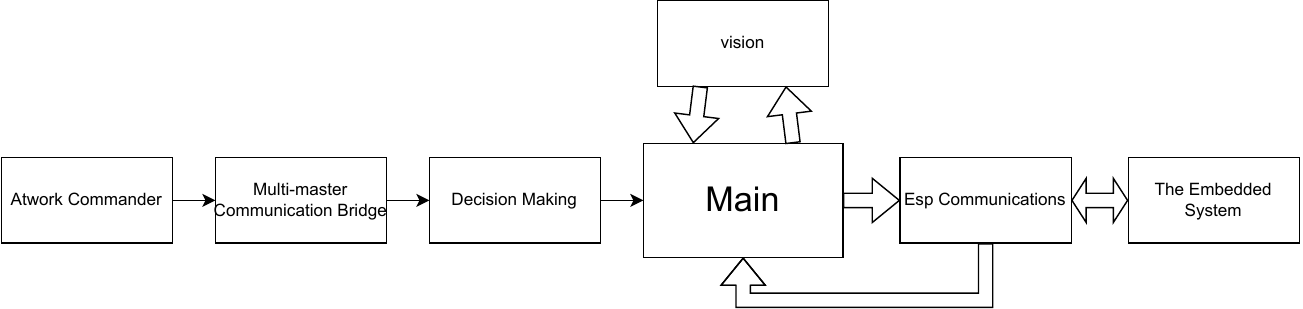}
    \caption{End-to-end data flow from task reception via atwork\_commander through decision-making, task execution, perception, and low-level actuation.}
    \label{fig:atwork_dataflow}
\end{figure}

Table~\ref{tab:atwork_ros_nodes} summarizes the nodes described in this section, and Table~\ref{tab:atwork_comm_channels} summarizes the distinct communication mechanisms used across the system. Figure~\ref{fig:atwork_rosgraph} shows the resulting node and topic graph, and Figure~\ref{fig:atwork_dataflow} illustrates the end-to-end data flow from task reception to actuation described throughout this section.

\begin{table}[H]
    \centering
    \small
    \caption{Summary of ROS nodes in the @Work Robot's software architecture}
    \renewcommand{\arraystretch}{1.3}
    \begin{tabularx}{\textwidth}{@{}llX@{}}
\toprule
\textbf{Node} & \textbf{Package} & \textbf{Primary Function} \\ \midrule
\texttt{decision\_making} & \texttt{decision\_making} & Receives the atwork\_commander task order and publishes an optimized, robot-specific task sequence (single-shot). \\
\texttt{main} & \texttt{navigationing} & Central finite-state-machine orchestrator: navigation, localization, and coordination of perception and actuation. \\
\texttt{esp\_comms} & \texttt{navigationing} & Serial bridge to the drivetrain/sensor microcontroller over UART; translates between ROS topics and the serial packet protocol. \\
\texttt{camera\_node} & \texttt{arm} & Acquires synchronized RGB-D frames from the RealSense D455 and exposes them via shared memory. \\
\texttt{vision\_node} & \texttt{arm} & Performs detection and LiDAR-based mapping; publishes results on \texttt{/vision\_out} and follows commanded state via the parameter server. \\
\bottomrule
\end{tabularx}
\label{tab:atwork_ros_nodes}
\end{table}

\begin{table}[H]
\centering
\small
\caption{Communication mechanisms used across the @Work Robot's software stack}
\renewcommand{\arraystretch}{1.3}
\begin{tabularx}{\textwidth}{@{}p{3.5cm}p{5.0cm}X@{}}
\toprule
\textbf{Mechanism} & \textbf{Used Between} & \textbf{Purpose} \\ \midrule
ROS topics (custom msgs) & \texttt{decision\_making} $\rightarrow$ \texttt{main} & One-shot transfer of the resolved task sequence (\texttt{decision}). \\
ROS topics (std\_msgs) & \texttt{main} $\leftrightarrow$ \texttt{esp\_comms} & Continuous actuation commands and sensor telemetry (\texttt{motor4}, \texttt{arm\_datas}, \texttt{encoder4}, \texttt{VL}, \texttt{cmps}). \\
ROS topics (std\_msgs) & \texttt{vision\_node} $\rightarrow$ \texttt{main} & Continuous perception output (\texttt{/vision\_out}). \\
ROS parameter server & \texttt{main} $\leftrightarrow$ \texttt{vision\_node} & Low-latency perception mode/target/pose commanding and state polling. \\
ROS parameter server & \texttt{esp\_comms} $\rightarrow$ \texttt{main} & Arm-response status flag. \\
POSIX shared memory & \texttt{camera\_node} $\rightarrow$ \texttt{vision\_node} & High-bandwidth RGB and depth frame transfer, bypassing ROS serialization. \\
Direct serial (non-ROS) & \texttt{vision\_node} $\leftrightarrow$ RPLidar & Direct sensor acquisition without an intermediate driver node. \\
Direct serial/UART (non-ROS) & \texttt{esp\_comms} $\leftrightarrow$ microcontroller & Fixed-format ASCII packet protocol carrying actuation commands and sensor telemetry. \\
\bottomrule
\end{tabularx}
\label{tab:atwork_comm_channels}
\end{table}

\section{Computer Vision and Perception}
\label{sec:cv}

\subsection{Mapping Subsystem}

The mapping subsystem pairs a LiDAR sensor with the Hector SLAM~\cite{hector_slam} algorithm to simultaneously estimate the robot's pose and build a two-dimensional map of the environment.

A key feature of Hector SLAM is that it doesn't depend on wheel odometry or IMU measurements, relying solely on laser scan data for localization and map construction. This makes the system robust to odometry drift and sensor noise.

The occupancy grid map is updated continuously across mission phases, including when it encounters static and dynamic obstacles. This keeps localization consistent and navigation reliable throughout task execution.

We chose Hector SLAM for its sensor independence, its accuracy in small- to medium-scale indoor environments, and its computational efficiency, which suits real-time operation on embedded hardware.

\subsection{Image Processing and Computer Vision}

The computer vision system is built around six primary operating modes, triggered by control signals from the robot's main module. It processes visual data and returns outputs for decision-making and task execution.

\subsubsection{Neural Network Development}

For real-time perception, we built a custom object detection and segmentation framework based on the YOLO~\cite{yolo} architecture, optimized for the target hardware.

We built a dedicated dataset to ensure robustness and generalization across varied environmental conditions. Images were captured directly from the robot's operational environment for realistic training scenarios, and synthetic images with varied backgrounds and lighting were generated to broaden generalization further. Pixel-level manual annotation was also carried out to support segmentation and improve detection accuracy.

\begin{figure}[H]
    \centering
    \includegraphics[width=0.85\linewidth]{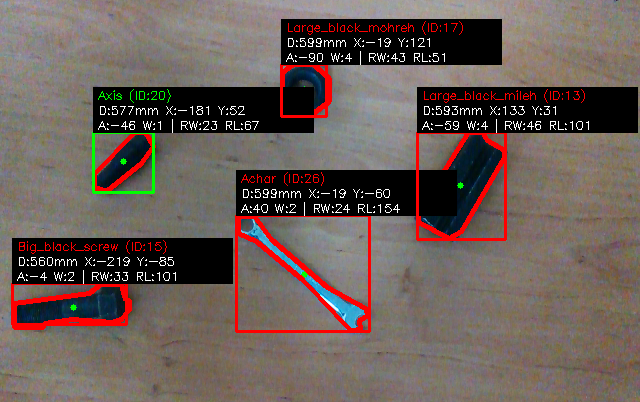}
    \caption{Real-time object detection and classification.}
    \label{fig:detection}
\end{figure}

Combining real, synthetic, and carefully annotated data gives the framework high detection precision and reliable segmentation, enabling real-time perception in dynamic environments. Figure~\ref{fig:detection} shows the system's real-time performance, with simultaneous detection, tracking, and physical dimension estimation of objects in the robot's surroundings. The values shown in the figure are the estimated real-world width (RW) and distance (D) from the sensor.

\paragraph{Final Dataset Statistics.}
The final dataset has 61,285 training images and 3,879 validation images, for 65,164 annotated samples in total. Roughly 6,000 of these (about 9\%) are original real-world images, with the remaining 59,164 generated through data augmentation for diversity. The dataset spans varied lighting conditions, object orientations, and backgrounds to improve robustness and generalization. All samples were manually checked for annotation consistency and segmentation accuracy.

\paragraph{Model Architecture and Training.}
\subparagraph{Base Model.}
YOLOv11s-Seg (small segmentation variant), chosen to balance accuracy against computational cost for real-time deployment.

\subparagraph{Task.}
Simultaneous object detection and instance segmentation, giving both bounding-box localization and pixel-level object mask prediction.

\subparagraph{Loss Functions.}
Training combines bounding box regression loss, classification loss, objectness loss, and segmentation loss to jointly optimize localization accuracy and mask quality.

\subparagraph{Training Strategy.}
The model was trained to convergence using performance-based early stopping, while monitoring mean Average Precision (mAP), precision, and recall to keep generalization stable.

\begin{figure}[H]
    \centering
    \includegraphics[width=\linewidth]{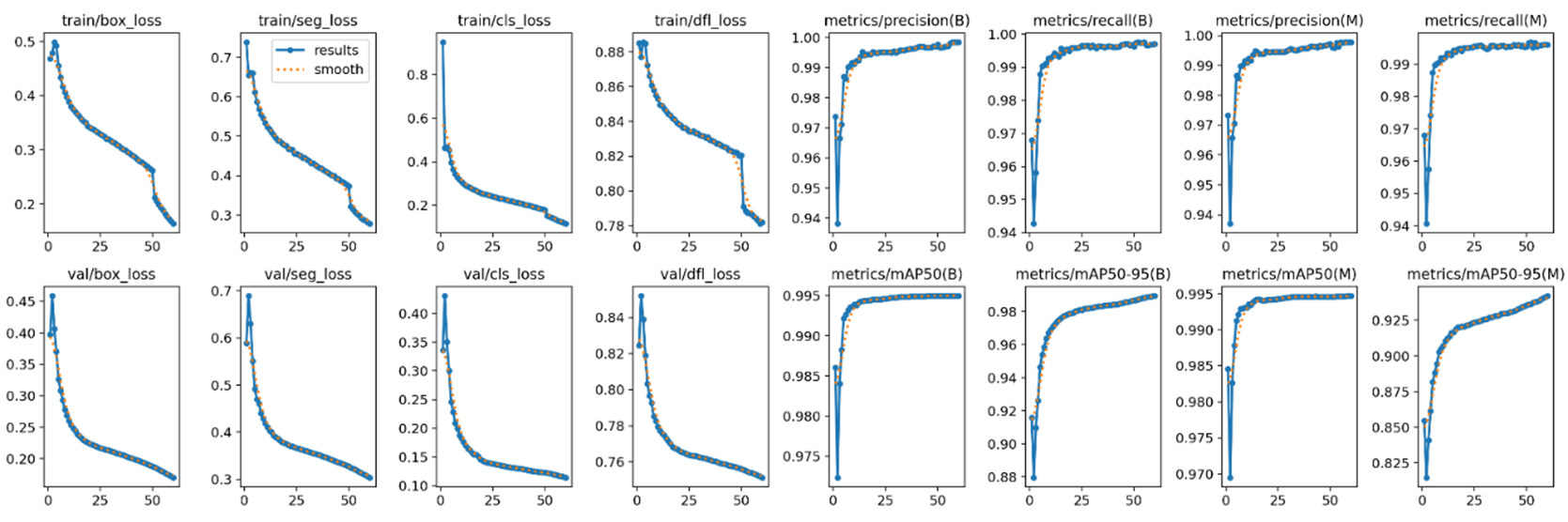}
    \caption{Training and validation loss curves, illustrating convergence across all loss components.}
    \label{fig:losscurves}
\end{figure}

\subparagraph{Experimental Results.}
The model trained for 50 epochs at a batch size of 32. The training and validation curves show stable convergence across all loss components — box, classification, segmentation, and distribution focal loss — as shown in Figure~\ref{fig:losscurves}. Both training and validation losses drop consistently with no sign of overfitting, which points to good generalization.

The model reaches a precision and recall of roughly 0.99 for both bounding box and segmentation outputs. The evaluation results also show an mAP50 near 0.995 and an mAP50-95 of about 0.98 for bounding boxes and 0.93 for segmentation masks.

The final deployed model runs at 17~FPS on the target hardware, confirming it's fast enough for real-time autonomous operation.

\subsubsection{Operational Modes}

\paragraph{Mode 1: Main Detection.}
This mode detects primary task-related objects using the YOLOv11s-Seg network, identifying both basic industrial components (e.g., bolts, nuts, rods) and more complex tools (e.g., drills, motors, bearings).

\subparagraph{Output Parameters.} By combining depth information with camera intrinsic parameters, the system computes, for each detected object: real-world dimensions (width/height), object orientation, the three-dimensional distance between the gripper and object center ($x, y, z$), and the gripper opening distance needed for a successful grasp.

\subparagraph{Post-Processing Classification.} Because some objects look visually similar, an additional dimension-based classification step assigns subcategories based on estimated physical measurements.

For example, a detected white rod is classified as a Large Rod if its estimated width exceeds 35~mm, and as a Small Rod otherwise. Spatial filtering constraints — such as a 1~meter maximum detection range — also help filter out irrelevant background detections.

\begin{figure}[H]
    \centering
    \includegraphics[width=0.7\linewidth]{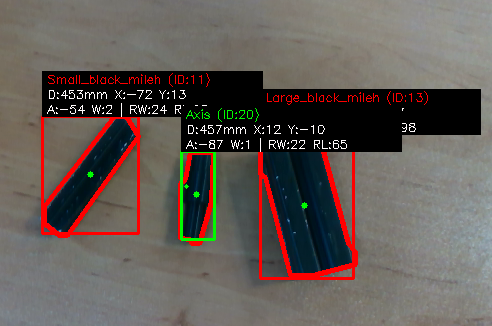}
    \caption{Subcategory classification based on estimated physical width (RW) and distance filtering (D).}
    \label{fig:subcategory}
\end{figure}

Figure~\ref{fig:subcategory} illustrates this post-processing step, where objects are sorted into subcategories (e.g., small vs. large) based on their RW and D as defined above.

Some objects start out under broader parent classes and get refined into specific child classes during post-processing. This hierarchical approach is needed because certain objects (e.g., large and small rods or nuts) look too similar for the network to reliably tell apart by appearance alone. Final categorization instead relies on estimated physical properties such as real-world dimensions and aspect ratio.

\paragraph{Mode 2: Precision Placement.}
This mode detects insertion points or empty target spaces using a lightweight YOLOv8n segmentation model, chosen for faster inference while keeping accuracy adequate.

\subparagraph{Geometric Analysis.} Since insertion points (e.g., rod slots and nut recesses) can look similar from a top-down view, appearance-based classification alone isn't enough. Detected shapes are first sorted into four geometric classes — Screw, Hexagon, Polygon, and Rectangle — and then refined based on aspect ratio and real-world dimensional thresholds.

\begin{figure}[H]
    \centering
    \includegraphics[width=0.6\linewidth]{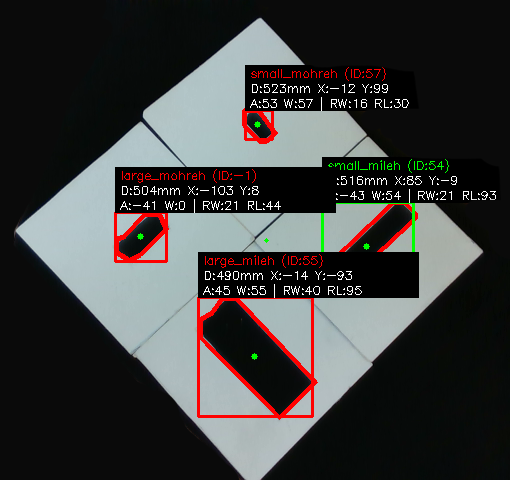}
    \caption{Geometric classification of objects based on real-world length (RL) and width (RW) thresholds for precision placement.}
    \label{fig:geometric_classification}
\end{figure}

\subparagraph{Classification Rules.} Objects are classified by their geometric characteristics. Rectangular shapes longer than 70~mm are identified as rod slots, and shorter ones as nut sockets. Nut sockets are further split by width: those under 35~mm are small nuts, and those above are large nuts. Figure~\ref{fig:geometric_classification} shows this classification process, with objects labeled small or large based on these real-world dimensional thresholds (RW and RL).

\paragraph{Mode 3: Navigation.}
This mode continuously updates the environment map with both virtual and physical obstacles, enabling dynamic path planning and safe navigation.

\subparagraph{A. Virtual Obstacles (Tape Detection).} Red and yellow tapes are detected with the YOLOv11s segmentation model. To account for the offset between the camera and LiDAR coordinate frames, a trigonometric transformation is applied: the horizontal distance is calculated with the Pythagorean theorem from the camera height and measured depth, then corrected for the camera--LiDAR spatial offset.

\begin{figure}[H]
    \centering
    \includegraphics[width=0.7\linewidth]{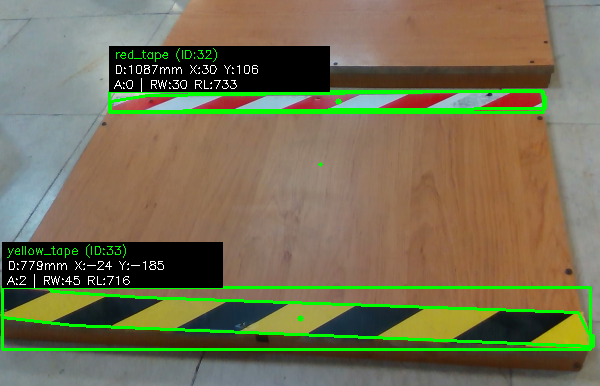}
    \caption{Segmentation and dimension estimation of virtual tape obstacles for global map projection.}
    \label{fig:tape}
\end{figure}

Each detected obstacle is then projected onto the global map with its estimated size and orientation, and a corresponding flag is sent to the control unit. As shown in Figure~\ref{fig:tape}, the system segments the tape obstacles and computes their real-world dimensions and distance, which are needed for accurate spatial projection and map updating.

\subparagraph{B. Physical Obstacles (LiDAR Detection).} Physical obstacles are found by comparing real-time LiDAR point cloud data against the original static map. The DBSCAN~\cite{dbscan} clustering algorithm groups nearby points and filters out noise, with the maximum distance between neighboring points (epsilon, $\varepsilon$) and the minimum cluster size (MinSamples) set accordingly.

\begin{figure}[H]
    \centering
    \includegraphics[width=0.5\linewidth]{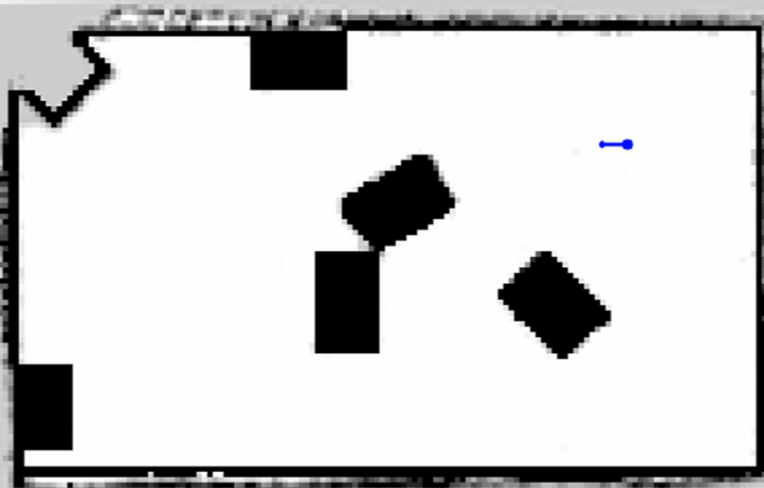}
    \caption{Real-time obstacle representation in the occupancy grid map based on LiDAR data and DBSCAN clustering.}
    \label{fig:lidar_clusters}
\end{figure}

Newly found clusters are flagged as dynamic obstacles and added to the updated map, and the control module is notified to trigger replanning. Figure~\ref{fig:lidar_clusters} shows the resulting local map, with LiDAR-detected clusters processed and represented as distinct physical obstacles for real-time navigation.

\paragraph{Mode 4: Zero Station and Physical Station.}
These modes align the robot with the center of a target station. Detection starts by extracting the front edge of the station with Canny edge detection, followed by line fitting using RANSAC (Random Sample Consensus)~\cite{ransac}.

We chose RANSAC because experiments showed it outperforms the traditional Hough Line Transform, with better robustness to noise and outliers while still reliably estimating the dominant line.

\begin{figure}[H]
    \centering

    \begin{subfigure}{0.47\textwidth}
        \centering
        \includegraphics[height=6cm]{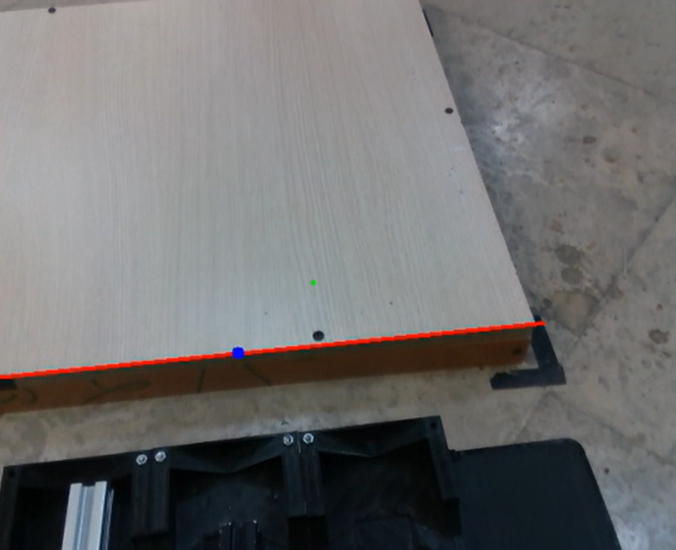}
        \caption{}
        \label{fig:14a}
    \end{subfigure}
    \hfill
    \begin{subfigure}{0.47\textwidth}
        \centering
        \includegraphics[height=6cm]{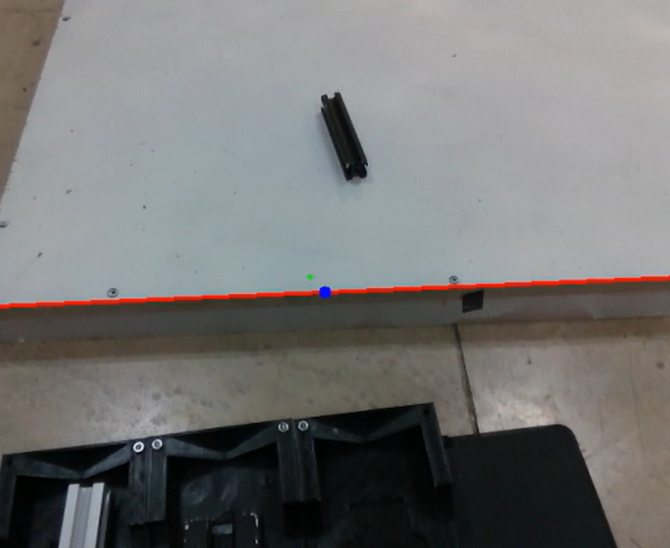}
        \caption{}
        \label{fig:14b}
    \end{subfigure}

    \caption{Robot alignment process using Canny edge detection and RANSAC line fitting: (a) detection of the station front edge, and (b) center point estimation for translational and angular correction.}
    \label{fig:ransac}
\end{figure}

From this, the system computes both the translational displacement needed for lateral alignment and the angular offset needed for orientation correction. Figure~\ref{fig:ransac} shows the edge detection and line fitting process on different stations, where the red line marks the estimated dominant edge and the blue point marks the calculated center for robot alignment.

\begin{figure}[H]
    \centering

    \begin{subfigure}{0.47\textwidth}
        \centering
        \includegraphics[height=6cm]{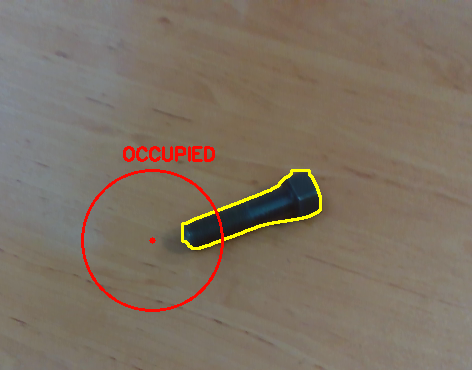}
        \caption{}
        \label{fig:15a}
    \end{subfigure}
    \hfill
    \begin{subfigure}{0.47\textwidth}
        \centering
        \includegraphics[height=6cm]{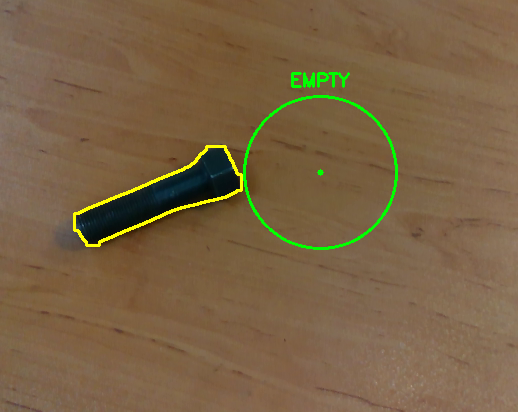}
        \caption{}
        \label{fig:15b}
    \end{subfigure}

    \caption{Emptiness check using a virtual scanning circle: (a) identification of an empty placement spot, and (b) detection of an occupied area.}
    \label{fig:emptiness}
\end{figure}

\paragraph{Mode 5: Emptiness Check.}
This mode checks whether a station has room for object placement. The system scans the station regions with a virtual circle of 5~cm radius, scaled by distance. When it finds an empty spot, the coordinates go to the control unit for the placement action that follows. Figure~\ref{fig:emptiness} shows this check, with a green circle marking a detected empty spot and a red circle marking an occupied region, preventing placement errors.

\paragraph{Mode 6: Container Detection.}
This mode detects containers and reports their center coordinates to the control unit. The red and blue container classes are part of the main YOLO Segmentation V11s model and are only detected when the control system explicitly requests it. Once a container is found, the system computes the real-world distance from the gripper to the container center (X, Y, Z) and publishes it.

\begin{figure}[H]
    \centering
    \includegraphics[width=0.8\linewidth]{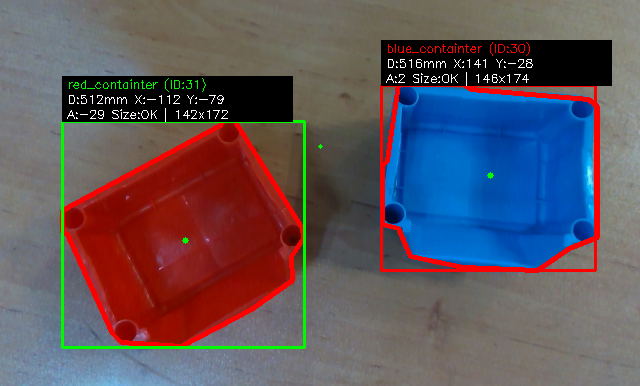}
    \caption{Real-time container detection and visibility verification for accurate center coordinate estimation.}
    \label{fig:container}
\end{figure}

To keep measurements reliable, data is only published once the entire container is visible in the camera's field of view, verified by width and height checks that rule out erroneous center calculations from partial views. Figure~\ref{fig:container} shows the detection of red and blue containers, where the system checks real-world dimensions (e.g., 146~$\times$~174~mm) to confirm the container is fully in view (labeled `Size:OK') before publishing coordinates.

\section{Control and Planning Architecture}
\label{sec:control}

\subsection{Localization and Mapping}

The map produced by the Computer Vision module is exported as a text file representing a discrete two-dimensional occupancy grid, stored for later use in navigation and decision-making.

Using this map together with the robot's kinematic model and onboard sensor measurements — gyroscope and wheel encoder data — the robot's global pose (position and orientation) is estimated in the world coordinate frame. This pose estimate and the occupancy grid then feed into path planning toward the target location.

The robot's kinematic model is described by the following equations~\cite{taheri_mecanum}:

\begin{equation}
\left\{
\begin{array}[c]{rl}
V_{xr} &= \dfrac{V_{FL} + V_{FR} + V_{BL} + V_{BR}}{4} \\[6pt]
V_{yr} &= \dfrac{-V_{FL} + V_{FR} + V_{BL} - V_{BR}}{4} \\[6pt]
\omega_z &= \dfrac{-V_{FL} + V_{FR} - V_{BL} + V_{BR}}{4}
\end{array}
\right.
\qquad
\left\{
\begin{array}[c]{rl}
V_x &= V_{xr} \cos\varphi \\[6pt]
V_y &= V_{yr} \sin\varphi \\[6pt]
\dot{\theta} &= \omega_z
\end{array}
\right.
\end{equation}

\begin{figure}[H]
    \centering
    \includegraphics[width=0.4\linewidth]{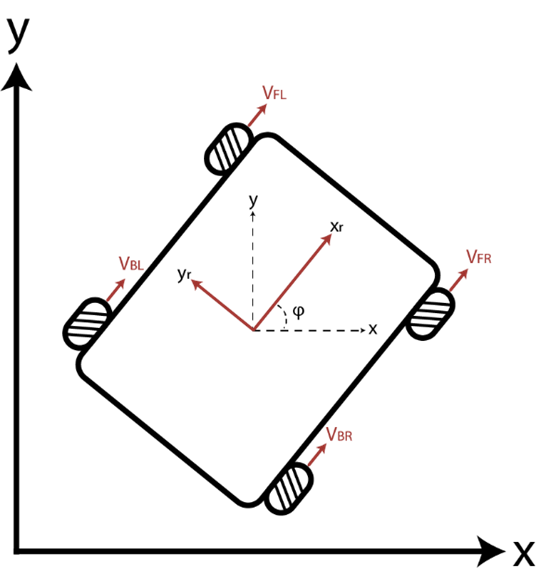}
    \caption{Kinematic representation of the mecanum-wheeled robot, showing the velocity vectors ($V_i$) for each wheel and the resulting motion vectors.}
    \label{fig:kinematics}
\end{figure}

This kinematic model, shown in Figure~\ref{fig:kinematics}, relates the individual wheel velocities to the robot's overall motion. For the four-wheel omnidirectional drive used here, global velocity breaks down into longitudinal, transverse, and angular components, letting the robot estimate its relative displacement (odometry) by integrating encoder data over time.

\subsection{Decision-Making and Path Planning}

The robot follows a structured decision-making and planning pipeline for task execution. Given a list of tasks, it first works out the optimal execution order with a dedicated decision-making algorithm, then a path planning algorithm computes an efficient route to each target location.

\subsubsection{Decision-Making}

The decision-making module evaluates all feasible orderings of task execution. Each candidate sequence is scored with a cost function that factors in path length and complexity, object type, target station type, number of empty slots, and the priority assigned to each task.

\begin{figure}[H]
    \centering
    \includegraphics[width=\linewidth]{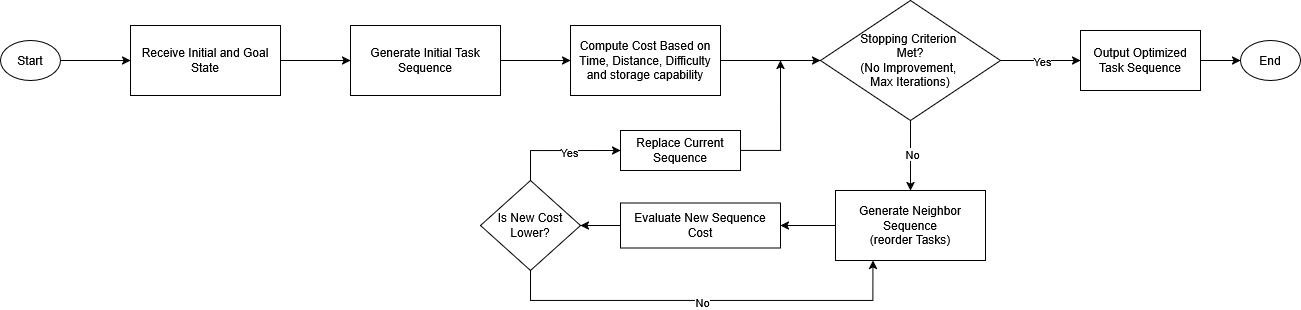}
    \caption{Flowchart of the task sequence optimization process, illustrating the iterative cost evaluation and neighbor sequence generation to achieve the most efficient execution plan.}
    \label{fig:decision_flowchart}
\end{figure}

Sequences that cost more than previously evaluated candidates are pruned along the way, so the system converges on the most cost-efficient task sequence, as shown in the optimization workflow in Figure~\ref{fig:decision_flowchart}.

The weights and coefficients for each cost parameter were tuned empirically through systematic experimentation to get robust decision-making performance.

\subsubsection{Path Planning}

The navigation system generates a collision-free trajectory from the robot's current pose to the target location given by the Decision Making module. Since the competition environment is a known occupancy grid, global path planning uses the A* search algorithm~\cite{astar}, giving an efficient, near-optimal path around static obstacles.
To speed things up~\cite{goyal_path}, A*'s open set is implemented as a Min-Heap priority queue, cutting the time complexity of selecting the next node from $O(n)$ to $O(\log n)$ — a real gain for longer paths in particular.
The node evaluation function is defined as follows.

\begin{equation} J=G+H+C_\text{Safety} \end{equation}

where the heuristic term is the Euclidean distance to the goal,

\begin{equation} H= \sqrt{(x-x_\text{goal})^2+(y-y_\text{goal})^2} \end{equation}
and the safety penalty is given by
$C_\text{Safety}=K/d_obs$  
where $d_\text{obs}$ is the distance to the nearest obstacle and $K$ is a tunable coefficient. This safety term biases the planner toward paths with more obstacle clearance, keeping the robot closer to the center of corridors and reducing collision risk from localization or tracking errors.
Once the optimal path is generated, it's passed to the motion controller as the reference trajectory. A PID-based path tracking controller continuously minimizes position and heading errors as the robot follows the planned route. During execution, LiDAR measurements are monitored in real time for unexpected obstacles; if a potential collision is detected, the robot stops immediately, updates the navigation map, replans the path, and resumes.
The whole navigation framework is implemented in C++ and runs in real time on the NVIDIA Jetson Orin NX, giving reliable, responsive autonomous navigation throughout the competition environment.
The overall system architecture follows the design described in \cite{mobile_manipulator}.

\subsection{Manipulator Structure and Object Handling Algorithm}

The robot's 5-degree-of-freedom (5-DOF) manipulator is controlled by an ESP32 microcontroller. High-level target poses and Cartesian coordinates are computed on the Jetson Orin NX and sent to the ESP32 for low-level execution.

Once it receives the target pose, the manipulator computes the corresponding joint angles from its inverse kinematic model. These joint commands are then applied to reach the end-effector position and orientation needed for object manipulation.

The forward kinematic model used to find the end-effector's position $(x,y,z)$ from the manipulator's geometry and joint angles is given by:

\begin{equation}        
\left\{
    \begin{array}[c]{rl}
        a &= \sqrt{l_1^2 + l_2^2 - 2l_1 l_2 \cos\beta} \\[6pt]
        b &= \sin^{-1}\!\left(\dfrac{l_2 \sin\beta}{a}\right) \\[6pt]
        c &= b\cos(\alpha - b)
    \end{array}
    \right.
    \qquad
    \left\{
        \begin{array}[c]{rl}
            x &= c\sin\phi \\[6pt]
            y &= c\cos\phi \\[6pt]
            z &= a\sin(\alpha - b)
        \end{array}
        \right.
\end{equation}

\begin{figure}[H]
    \centering
    \includegraphics[width=0.4\linewidth]{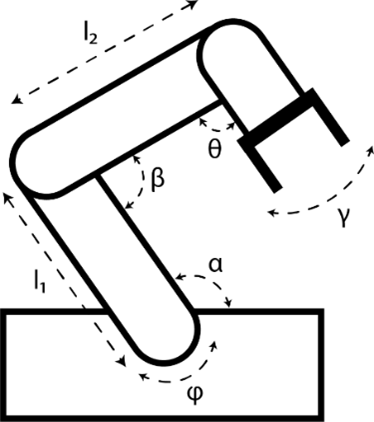}
    \caption{Geometric model of the manipulator highlighting the joint angles and link structure used for kinematic calculations.}
    \label{fig:manipulator_geometry}
\end{figure}

Here $l_1$ and $l_2$ are the lengths of the manipulator links, and $\alpha$, $\beta$, $\phi$ are the respective joint angles, as shown in Figure~\ref{fig:manipulator_geometry}. These equations let the high-level controller map the manipulator's internal state to the operational workspace coordinates.

\subsubsection{Object Placement Slots}

As mentioned earlier, the robot has a set of object placement slots at its front. These slots move linearly along the horizontal axis (left and right). During pickup and placement, a predefined control algorithm positions the slots so the target slot sits directly in front of the manipulator.

Since the system always knows the exact position of each slot through this precise control, no extra scanning, perception, or processing is needed to find slot positions during pickup or placement. This cuts computational overhead and makes object-handling more efficient and reliable.

\subsubsection{Object Picking}

At the start of the picking procedure, the robot first checks it has enough storage capacity for a new object. If there's room, the manipulator moves to a predefined hovering position above the target station.

Once it receives the precise pose of the target object from the perception module, the manipulator approaches and grasps it with the end-effector gripper. Real-time load-cell measurements then verify the grasp; once confirmed, the object is lifted and moved to the robot's front storage compartment, where it stays secured until delivery.

\subsubsection{Object Placement}

During placement, the manipulator retrieves the object from the front storage compartment and begins the placement procedure. The system first identifies the available section of the target station for placement.

Once a free slot is found, visual data from the perception module gives the precise coordinates of the target location. From this, the manipulator computes the required joint configuration and aligns the end-effector with the placement area.

Finally, the gripper releases the object in a controlled manner, completing the placement.

\subsubsection{Performance Evaluation}

To keep the placement procedure reliable, we set specific operational constraints and performance metrics:

\textbf{Operational Speed:} Each manipulator joint can reach a maximum angular velocity of $40^{\circ}/\mathrm{s}$. To keep motion controllable and stable and to ensure safe operation in the lab, though, joint velocities are capped between $15^{\circ}/\mathrm{s}$ and $20^{\circ}/\mathrm{s}$, depending on conditions.

\textbf{Grasping Success Rate:} We ran a series of grasping experiments on objects of various shapes, sizes, and surface properties to evaluate the manipulator's performance, testing each object over 20 independent trials. A trial counted as successful if the robot could grasp, lift, and place the object at the target location without dropping it. Table~\ref{tab:grasping} summarizes the resulting success rates.

\begin{table}[H]
\centering
\caption{Experimental results of grasping and placement success rates for various industrial objects, categorized by dimensions, weight, and geometric complexity.}
\label{tab:grasping}
\renewcommand{\arraystretch}{1.15}
\begin{tabular}{@{}lccccc@{}}
\toprule
\textbf{Object} & \textbf{Dimensions} & \textbf{Weight (g)} & \textbf{Trials} & \textbf{Successes} & \textbf{Success Rate (\%)} \\ \midrule
Small Aluminum Profile & 20 $\times$ 20 $\times$ 100 mm & 50  & 20 & 19 & 95 \\
Large Aluminum Profile & 40 $\times$ 40 $\times$ 100 mm & 185 & 20 & 18 & 90 \\
Screw                  & M20 $\times$ 100 mm            & 295 & 20 & 16 & 80 \\
Small Nut              & M20                             & 55  & 20 & 9  & 45 \\
Large Nut              & M30                             & 215 & 20 & 11 & 55 \\
Steel Axis             & 68 $\times$ M20                 & 160 & 20 & 15 & 75 \\
Bearing                & 40 $\times$ 17                  & 180 & 20 & 8  & 40 \\
Motor 755              & 66.7 $\times$ 42                & 320 & 20 & 15 & 75 \\
Wrench                 & 144 $\times$ 20                 & 75  & 20 & 9  & 45 \\
Wera Allen Key 8 mm    & 195 $\times$ 37 $\times$ 8 mm    & 15  & 20 & 7  & 35 \\
Drill                  & 150 $\times$ 13 $\times$ 13 mm   & 17  & 20 & 10 & 50 \\
\bottomrule
\end{tabular}
\end{table}

\newpage

\begin{center}
    \large\textbf{Author Contributions}
    
    \vspace{0.75em}
    \small
    \renewcommand{\arraystretch}{1.25}
    \begin{tabularx}{\textwidth}{|>{\raggedright\arraybackslash\bfseries}p{4.3cm}|>{\raggedright\arraybackslash}X|}
    \hline
    \textbf{Name(s)} & \textbf{Role and Contribution Summary} \\
    \hline
    
    Aria Dadnavi &
    Developed the post-processing pipeline, adaptive LiDAR--camera mapping system, and computer vision algorithms for object detection and pick-and-place perception. \\
    \hline

    Matin Jamshidi &
    Mechanical design and integration of the robot chassis, manipulator arms, and gripper mechanism, including fabrication, assembly, and structural optimization. \\
    \hline

    Kian Khatibi &
    Embedded firmware design, real-time sensor processing, actuator interfacing, and communication system implementation. \\
    \hline

    Shahriar Karimi Makhsous &
    ROS architecture and system communication design. \\
    \hline

    Saina Najafi Shabestari &
    Trained and optimized the instance segmentation model and implemented object detection algorithms for pick-and-place perception. \\
    \hline

    Seyed Sobhan Hosseini Hajibekandeh,
    Amirhossein Gholizadeh Behbahani &
    Developed the autonomous navigation, motion planning, manipulator control, and task execution algorithms. \\
    \hline

    Zahra Maham,
    Reyhane Mohasseli &
    Electronics hardware design, sensor integration, PCB development, and power management. \\
    \hline

    Ghazal Ghoreishi &
    Implemented computer vision algorithms for station alignment, robot motion correction, and pick-and-place perception. \\
    \hline

    Mohammad Hossein Moaiyeri &
    Team supervision, university liaison, and funding coordination. \\
    
    \hline
    \end{tabularx}
    \end{center}

\section*{Acknowledgments}
The authors wish to note that the submission of this manuscript was delayed due to prolonged disruptions to research activities caused by civil unrest and subsequent military conflict in Iran. These circumstances significantly restricted access to laboratory facilities and delayed manuscript preparation. The work presented here corresponds to the robotic system as completed in early 2025.

Additionally, the authors used AI language models to assist with English-language editing, \LaTeX{} formatting, and manuscript preparation, owing to limited fluency in English. All technical content, experiments, data, and conclusions are the authors' own, and the authors have reviewed the text and take full responsibility for its contents.

\end{document}